\documentclass[letterpaper, 10 pt, conference]{ieeeconf} 
\usepackage[LGR,T1]{fontenc}
\usepackage[latin9]{inputenc}
\usepackage{verbatim}
\usepackage{amsmath}
\usepackage{amssymb}
\usepackage{graphicx}
\usepackage{graphics}
\usepackage{epsfig}
\usepackage{epstopdf}
\usepackage[caption=false]{subfig}
\usepackage{cite}

\newtheorem{rem}{Remark}
\newtheorem{problem}{Problem}
\newtheorem{thm}{Theorem}
\newtheorem{defn}{Definition}

\IEEEoverridecommandlockouts
\title{Decentralized Motion Planning with Collision Avoidance for a Team of UAVs under High Level Goals}

\author{Christos K. Verginis, Ziwei Xu and Dimos V. Dimarogonas \thanks{The authors are with the Centre for Autonomous Systems and ACCESS Linnaeus Centre, KTH Royal Institute of Technology, Stockholm 10044, Sweden. Emails: \{cverginis, dimos, ziwei\}@kth.se.} \thanks{This work was supported by funding from the Knut and Alice Wallenberg Foundation, the Swedish Research Council (VR), the European Union's Horizon 2020 Research and Innovation Programme under the Grant Agreement No. 644128 (AEROWORKS) and the H2020 ERC Starting Grant BUCOPHSYS. } }

\begin{document}

\maketitle

\begin{abstract}
This paper addresses the motion planning problem for a team of aerial agents under high level goals. We propose a hybrid control strategy that guarantees the accomplishment of each agent's local goal specification, which is given as a temporal logic formula, while guaranteeing inter-agent collision avoidance. In particular, by defining $3$-D spheres that bound the agents' volume, we extend previous work on decentralized navigation functions and propose control laws that navigate the agents among predefined regions of interest of the workspace while avoiding collision with each other. This allows us to abstract the motion of the agents as finite transition systems and, by employing standard formal verification techniques, to derive a high-level control algorithm that satisfies the agents' specifications. Simulation and experimental results with quadrotors verify the validity of the proposed method.
\end{abstract}

\section{Introduction}
Multi-agent systems have received a significant amount of attention over the last decades. The complexity of many tasks, such as assembling parts and heavy/large object transportation or manipulation, necessitates the employment of a group of robots, rather than a single one, since it offers greater versatility, redundancy and fault tolerance. 

In the case of aerial vehicles, tasks involving area coverage/inspection or rescue missions point out the importance of using multi-agent setups. The standard problem of formation control for a team of aerial vehicles is addressed in \cite{Liu_ICIA2015,Yu_CCC2013, Koksal_ECC2015, Hou_ECC2016, Pereira_MED2016, Dong_ICST2015}, whereas \cite{Mercado_ECC2013,Roldao_ECC2013,Ulrigh_ACC2016,Hao_ACC2016,Ghamry_ICUAS2015} consider leader-follower formation approaches, where the latter also treats the problem of collision avoidance with static obstacles in the environment; \cite{Sunberg_ICRA2016}, \cite{Eskandarpour_ICROM2014} and \cite{Zhou_CDC2015} employ dynamic programming, Model Predictive Control and reachable set algorithms, respectively, for inter-agent collision avoidance. In \cite{Pierson_ICRA2016} the cooperative evader pursuit problem is treated.

Ultimately, however, we would like the aerial robots to execute more complex high-level tasks, involving combinations of safety ("never enter a dangerous regions"), surveillance ("keep visiting regions $A$ and $B$ infinitely often") or sequencing ("collect data in region $C$ and upload it in region $D$") properties. 
Temporal logic languages offer a means to express the aforementioned specifications, since they can describe complex planning objectives in a more efficient way than the well-studied navigation algorithms. Recently, the incorporation of temporal logic-based planning to the robotics and automation field has gained a considerable amount of attention, both in single- and multi-agent setups \cite{Quottrup_ICRA2004,Loizou_CDC2005, Filippidis_CDC2012, Guo_IJRR2015,Nenchev_ECC2016,Feyzabadi_ICRA2016,Guo_CDC2015,Fainekos_Automatica2009,Kloetzer_ICNSC2006,Ulusoy_IJRR2013,Zhang_ACC2016, Aksaray_ICRA16}. Regarding aerial vehicles, \cite{Karaman_CDC2008} addresses the vehicle routing problem using MTL specifications and \cite{Karaman_ACC2008} approaches the LTL motion planning using MILP optimization techniques, both in a centralized manner. Markov Decision Processes are used for the LTL planning in \cite{Xiaoting_CCC2016}. The aforementioned works, however, consider discrete agent models and do not take into account their continuous dynamics. 

Another important feature in multi-agent planning and control is the need for decentralization and minimization of inter-agent communication; centralized approaches, where a central system computes the overall team plan or cases where the agents communicate online with each other, can cause computationally expensive procedures, even for small robot teams. In the case of temporal logics, the use of product transition systems incorporating the states of all agents (as e.g., in \cite{Ulusoy_IJRR2013,Loizou_CDC2005,Xiaoting_CCC2016}) can render the solution to the motion planning problem practically infeasible. 

Moreover, the majority of works in the related literature of temporal logic-based motion planning considers point-agents (as, e.g. in \cite{Guo_IJRR2015,Kloetzer_ICNSC2006,Fainekos_Automatica2009}) and does not take into account potential collisions between them. The latter is a crucial safety property in real-time scenarios, where actual vehicles are used in the motion planning framework.

In this work, we propose a novel decentralized control protocol for the motion planning of a team of aerial vehicles under LTL specifications with simultaneous inter-agent collision avoidance. In particular, we extend previous work on decentralized navigation functions \cite{Dimos_2007_journal} to abstract the motion of each agent as a finite transition system. Then, we employ standard formal-verification techniques to derive plans that satisfy each agent's LTL specification. The proposed control protocol is decentralized in the sense that each agent has limited sensing information and derives and executes its desired path without communicating with the other agents or knowing their respective goals/specifications. Simulation and experimental results with quadrotors verify the effectiveness of the proposed framework. To the best of the authors' knowledge, this is the first approach that integrates temporal logic-based motion planning with decentralized navigation functions in an experimental framework with UAVs.  

The rest of the paper is organized as follows: Section \ref{sec:Notation-and-Preliminaries} introduces notation and preliminary background. Section \ref{sec:System and PF} describes the problem formulation and the overall system's model. The control strategy is presented in Section \ref{sec:control strategy}. Sections \ref{sec:simulation results} and \ref{sec:experimental results} verify our approach through numerical simulations and experiments, respectively, and Section \ref{sec:conclusion} concludes the paper.

\section{Notation and Preliminaries\label{sec:Notation-and-Preliminaries}}

Vectors and matrices are denoted with bold lowercase and uppercase letters, respectively, whereas scalars are denoted with non-bold letters. 
The set of positive integers is denoted as $\mathbb{N}$ and the real $n$-coordinate space, with $n\in\mathbb{N}$, as $\mathbb{R}^n$; $\mathbb{R}^n_{\geq 0}$ is the set of real $n$-vectors will all elements nonnegative;
$\mathcal{B}_r(\boldsymbol{c})$ denotes the ball of radius $r \geq 0$ and center $\boldsymbol{c}\in\mathbb{R}^{3}$; Moreover, given a set $A$, the notation $\mathring{A}$ denotes the interior of $A$, $2^A$ is the set of all subsets of $A$ and, given a finite sequence $a_1,\dots,a_n$ of elements in $A$, with $n\in\mathbb{N}$, we denote by $(a_1,\dots,a_n)^\omega$ the infinite sequence $a_1,\dots,a_na_1,\dots,a_n\dots$ created by repeating $a_1,\dots,a_n$. Finally, $d_n:\mathbb{R}^n\times\mathbb{R}^n \rightarrow \mathbb{R}_{\geq 0}$ is the $n$-dimensional Euclidean distance, with $n\in\mathbb{N}$.

\subsection{Specification in LTL \label{subsec:LTL}} 

We focus on the task specification $\phi$ given as a Linear Temporal Logic (LTL) formula. The basic ingredients of a LTL formula are a set of atomic propositions $\mathcal{AP}$ and several boolean and temporal operators. LTL formulas are formed accoding to the following grammar \cite{BayeKatoen_2008}: $\phi ::= \mathsf{true}\: |\:a\: |\: \phi_{1} \land  \phi_{2}\: |\: \neg \phi\: |\:\bigcirc \phi\:|\:\phi_{1}\cup\phi_{2} $, where $a\in\mathcal{AP}$ and $\bigcirc$ (next), $\cup$ (until). Definitions of other useful operators like $\square$ (\it always\rm), $\lozenge$ (\it eventually\rm) and $\Rightarrow$ (\it implication\rm) are omitted and can be found in \cite{BayeKatoen_2008}.

The semantics of LTL are defined over infinite words over $2^{\mathcal{AP}}$. Intuitively, an atomic proposition $\psi\in\mathcal{AP}$ is satisfied on a word $w=w_1w_2\dots$ if it holds at its first position $w_1$, i.e. $\psi\in w_1$. Formula $\bigcirc\phi$ holds true if $\phi$ is satisfied on the word suffix that begins in the next position $w_2$, whereas $\phi_1\cup\phi_2$ states that $\phi_1$ has to be true until $\phi_2$ becomes true. Finally, $\lozenge\phi$ and  $\square\phi$ holds on $w$ eventually and always, respectively. For a full formal definition of the LTL semantics, the reader is kindly referred to \cite{BayeKatoen_2008}.
 
\subsection{Navigation Functions \label{subsec:NF}}

Navigation functions, first introduced in \cite{Koditchek_1990}, are real valued maps realized through cost functions, whose negated gradient field is attractive towards the goal configuration and repulsive with respect to obstacles. A navigation function can be formally defined as follows:  
\begin{defn}
Let $F\subset R^n$ be a compact connected analytic manifold with boundary. A map $\phi:F\rightarrow \left[0,1\right]$ is a navigation function if: (1) It is analytic on $F$, (2) it has only one minimum $\boldsymbol{q_d}\in \mathring{F}$, (3) its Hessian at all critical points (zero gradient field) is full rank and (4), $\lim\limits_{\boldsymbol{q}\rightarrow\boldsymbol{\partial F}}\phi(\boldsymbol{q})=1$.
\end{defn}
Following \cite{Koditchek_1990}, given a spherical workspace $F$ centered at $\boldsymbol{q_0}\in F$ with radius $r_0\geq 0$, an initial position $\boldsymbol{q_s}\in\mathring{F}$,a goal position $\boldsymbol{q_d}\in\mathring{F}$ and $M$ spherical obstacles with center and radius $\boldsymbol{q_j}\in F$, $r_j\geq 0$ respectively for $j=1,\cdots,M$, a choice for a navigation function in $F$ is $\Phi:F\rightarrow[0,1]$, with:
\begin{equation}
\Phi(\boldsymbol{q}(t)) = \dfrac{\gamma(\boldsymbol{q})}{(\gamma^k(\boldsymbol{q}) + \beta(\boldsymbol{q}))^{1/k}},
\end{equation}
where $k > 0$ is a design parameter, $\boldsymbol{q}:\mathbb{R}_{\geq 0}\rightarrow\mathring{F}, \gamma:\mathring{F}\rightarrow\mathbb{R}_{\geq 0}$, with $\gamma(\boldsymbol{q})= \lVert \boldsymbol{q} - \boldsymbol{q_d} \rVert ^2$, is the attractive potential towards the goal and $\beta:\mathring{F}\rightarrow\mathbb{R}$, with $\beta(\boldsymbol{q}) = \prod_{j=0}^M \beta_j(\boldsymbol{q})$, is the repulsive potential from the workspace boundary and the obstacles, where $\beta_0(\boldsymbol{q})=r_0^2 - \lVert \boldsymbol{q} - \boldsymbol{q_0} \rVert^2$ and $\beta_j(\boldsymbol{q}) = \lVert \boldsymbol{q} - \boldsymbol{q_j} \rVert^2 - r_j^2, j =1,\cdots,M$;  $\Phi(\boldsymbol{q})$ reaches its minimal value $0$ only at $\boldsymbol{q_d}$ and its maximal value $1$ at the boundaries of the workspace and the obstacles. It has been shown that by following the negated gradient $-\nabla_{\boldsymbol{q}}\Phi$, it is guaranteed for sufficiently large $k$ that $\lim_{t\rightarrow\infty}\gamma(\boldsymbol{q}(t)) = 0$ and $\beta(\boldsymbol{q}(t)) > 0, \forall t\geq 0$, for almost all initial positions $\boldsymbol{q_s}\in\mathring{F}$.

\section{System and Problem Formulation} \label{sec:System and PF}
Consider $N$ aerial agents operating in a static workspace that is bounded by a large sphere in $3$-D space $\mathcal{W}=\mathcal{B}_{r_0}(\boldsymbol{p_0})=\{\boldsymbol{p}\in \mathbb{R}^3\text{ s.t. } \lVert \boldsymbol{p}-\boldsymbol{p_0} \rVert\leq r_0 \}$, where $\boldsymbol{p_0}\in \mathbb{R}^3$ and $r_0>0$ are the center and radius of $\mathcal{W}$. Within $\mathcal{W}$ there exist $K$ smaller spheres around points of interest, which are described by $\mathcal{\pi}_k=\mathcal{B}_{r_{\pi_k}}(\boldsymbol{p_{\pi_k}})=\{\boldsymbol{p}\in \mathbb{R}^3\text{ s.t. } \lVert \boldsymbol{p}-\boldsymbol{p_{\pi_k}} \rVert\leq r_{\pi_k} \}\subset \mathcal{W}$, where $\boldsymbol{p_{\pi_k}}\in \mathbb{R}^3, r_{\pi_k}>0$ are the central point and radius, respectively, of $\pi_k$. We denote the set of all $\pi_k$ as $\Pi=\{\pi_1,\dots,\pi_K \}$. For the workspace partition to be valid, we consider that the regions of interest are sufficiently distant from each other and from the workspace boundary, i.e., $d_3(\boldsymbol{p_{\pi_k}},\boldsymbol{p_{\pi_{k'}}}) > 4\max_{k\in\{1,\dots,K\}}(r_{\pi_k})$ and $d_3(\boldsymbol{p_{\pi_k}},\boldsymbol{p_0})< r_0-3r_{\pi_k}, \forall k,k'\in\{1,\dots,K\}$ with $k\neq k'$. Moreover, we introduce a set of atomic propositions $\Psi_i$ for each agent $i\in\{1,\dots,N\}$ that indicates certain properties of interest of agent $i$  in $\Pi$ and are expressed as boolean variables. The properties satisfied at each region $\pi_k$ are provided by the labeling function $\mathcal{L}_i:\Pi\rightarrow 2^{\Psi_i}$, which assigns to each region $\pi_k, k\in\{1,\dots,K\}$ the subset of the atomic propositions $\Psi_i$ that are true in that region.  
  
\begin{figure}[!btp]
\centering
\includegraphics[trim = 0cm 0.5cm 0cm -0.5cm,width = 0.2\textwidth, height = 0.13\textheight]{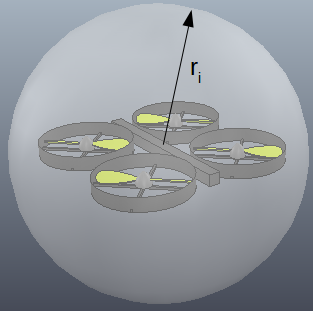}
  
\caption{Bounding sphere of an aerial vehicle.\label{fig:quad}}
\end{figure} 

\subsection{System model \label{subsec:system_model}}
Each agent $i\in\{1,\dots,N\}$ occupies a bounding sphere: $\mathcal{B}_{r_i}(\boldsymbol{p_i}(t)) = \left\{\boldsymbol{p}(t)\in\mathcal{W} \text{ s.t. } \lVert \boldsymbol{p}(t)-\boldsymbol{p_i}(t) \rVert \leq r_i \right\}$, where $\boldsymbol{p_i}:\mathbb{R}_{\geq 0}\rightarrow \mathbb{R}^3$ is the center and $r_i > 0$ the radius of the sphere (Fig. \ref{fig:quad}). We also consider that $r_i < r_{\pi_k}, \forall i\in\{1,\dots,N\},k\in\{1,\dots,K\}$, i.e., the regions of interest are larger than the aerial vehicles. The motion of each agent is controlled via its centroid $\boldsymbol{p_i}$ through the single integrator dynamics: 
\begin{equation}
\boldsymbol{\dot{p}_i} = \boldsymbol{u_i}, i\in\{1,\dots,N\}. \label{eq:dynamics}
\end{equation}
Moreover, we consider that agent $i$ has a limited sensing range of $d_{s_i} > \max_{i,j=\{1,\dots,N\}}(r_i+r_j)$. Therefore, by defining the neighboring set $\mathcal{N}_i = \{ j\in\{1,\dots,N\}, \text{ s.t. } \Vert \boldsymbol{p}_i - \boldsymbol{p}_j \rVert \leq d_{s_i} \}$, agent $i$ knows at each time instant the position of all $\boldsymbol{p_j}, \forall j\in\mathcal{N}_i$ as well as its own position $\boldsymbol{p}_i$. The workspace is perfectly known, i.e., $\boldsymbol{p_{\pi_k}},r_{\pi_k}$ are known to all agents, for all $k\in\{1,\dots,K\}$. 

With the above ingredients, we provide the following definitions:
\begin{defn} \label{def:agent in region}
An agent $i\in\{1,\dots,N\}$ is in a region $\pi_k,k\in\{1,\dots,K\}$ at a configuration $\boldsymbol{p_i}$, denoted as $\mathcal{A}_i(\boldsymbol{p_i})\in\pi_k$, if and only if $\mathcal{B}_{r_i}(\boldsymbol{p_i}) \subseteq \mathcal{B}_{r_{\pi_k}}(\boldsymbol{p_{\pi_k}})$.
\end{defn}
\begin{defn} \label{def:agent_transition}
Assume that $\mathcal{A}_i(\boldsymbol{p_i}(t_0))\in\pi_k, i\in\{1,\dots,N\},k\in\{1,\dots,K\}$ for some $t_0\geq 0$. Then there exists a transition for agent $i$ from region $\pi_k$ to region $\pi_{k'},k'\in\{1,\dots,K\}$, denoted as $\pi_k\rightarrow_i\pi_{k'}$, if and only if there exists a finite $t_f\geq 0$ and a bounded control trajectory $\boldsymbol{u_i}$ such that (i) $\mathcal{A}_i(\boldsymbol{p_i}(t_f))\in\pi_{k'}$, (ii) $\mathcal{B}_{r_i}(\boldsymbol{p_i}(t))\cap\mathcal{B}_{r_{\pi_m}}(\boldsymbol{p_{\pi_m}}) = \emptyset$, and (iii) $\mathcal{B}_{r_i}(\boldsymbol{p_i}(t))\cap\mathcal{B}_{r_{i'}}(\boldsymbol{p_{i'}}(t)) = \emptyset, \forall m\in\{1,\dots,K\}$ with $m\neq k,k', \forall i'\in\{1,\dots,N\}$ with $i'\neq i$ and $t\in\left[0,t_f \right]$.
\end{defn}

Loosely speaking, an agent $i$ can transit between two regions of interest $\pi_k$ and $\pi_{k'}$, if there exists a bounded control trajectory $\boldsymbol{u_i}$ in (\ref{eq:dynamics}) that takes agent $i$ from $\pi_k$ to $\pi_{k'}$ while avoiding entering all other regions and colliding with the other agents.

\subsection{Specification}
Our goal is to control the multi-agent system subject to (\ref{eq:dynamics}) so that each agent's behavior obeys a given specification over its atomic propositions $\Psi_i$. 

Given a trajectory $\boldsymbol{p_i}(t)$ of agent $i$, its corresponding \textit{behavior} is given by the infinite sequence $\beta_i = (\boldsymbol{p_i}(t),\psi_i)=(\boldsymbol{p_{i_1}},\psi_{i_1})(\boldsymbol{p_{i_2}},\psi_{i_2})\dots$, with $\psi_{i_m}\in2^{\Psi_i}$ and $\mathcal{A}(\boldsymbol{p_{i_m}})\in\pi_{k_m},  \psi_{i_m}\in\mathcal{L}_i(\pi_{k_m}),k_m\in\{1,\dots,K\},\forall m\in\mathbb{N}$. 

\begin{defn}
The behavior $\beta_i = (\boldsymbol{p_i}(t),\psi_i)$ satisfies an LTL formula $\phi$ if and only if $\psi_i \models \phi$.
\end{defn}

\subsection{Problem Formulation}
The control objectives are given for each agent separately as LTL formulas $\phi_i$ over $\Psi_i, i\in\{1,\dots,N\}$. An LTL formula is satisfied if there exists a behavior $\beta_i = (\boldsymbol{p_i}(t),\psi_i)$ of agent $i$ that satisfies $\phi_i$. Formally, the problem treated in this paper is the following:
\begin{problem} \label{problem}
Given a set of aerial vehicles $N$ subject to the dynamics (\ref{eq:dynamics}) and $N$ LTL formulas $\phi_i,$ over the respective atomic propositions $\Psi_i, i\in\{1,\dots,N\}$, achieve behaviors $\beta_i$ that (i) yield satisfaction of $\phi_i,\forall i\in\{1,\dots,N\}$ and (ii) guarantee inter-agent collision avoidance.
\end{problem}

\section{Main Results} \label{sec:control strategy}

\subsection{Continuous Control Design} \label{subsec:Continuous Control}

The first ingredient of our solution is the development of a decentralized feedback control law that establishes a transition relation $\pi_k\rightarrow_i\pi_{k'}, \forall k,k'\in\{1,\dots,K\}$ according to Def. \ref{def:agent_transition}. First, we provide an overview of the concept of \textit{Decentralized Navigation Functions}, introduced in \cite{Dimos_2007_journal}, that we will use in the subsequent analysis. 

\subsubsection{\textbf{Decentralized Navigation Functions (DNFs)}} 
Consider $N$ agents described by the position variables $\boldsymbol{p_i}(t)\in \mathbb{R}^3$, bounding spheres $\mathcal{B}_{r_i}(\boldsymbol{p_i}(t))$, sensing radius $d_{s_i}>0, i\in\{1,\dots,N\}$ and dynamics as in (\ref{eq:dynamics}). Each agent's goal is to reach a desired position $\boldsymbol{p^d_i}\in \mathbb{R}^3$ without colliding with the other agents. To this end, we employ the following class of \textit{decentralized navigation functions}: $\phi_i:\mathbb{R}^{3N}\rightarrow[0,1]$, with $\phi_i(\boldsymbol{p}) = \dfrac{\gamma_i(\boldsymbol{p_i}) + f_i(G_i)}{ (\gamma_i(\boldsymbol{p_i})^{\lambda_i} + G_i(\boldsymbol{p}))^{1/\lambda_i} }$, where $\boldsymbol{p} = [\boldsymbol{p_1},\dots,\boldsymbol{p_N}]^T$ and $\lambda_i > 0$. The function $\gamma_i:\mathbb{R}^3\rightarrow\mathbb{R}_{\geq 0}$ is defined as $\gamma_i(\boldsymbol{p_i}) = \lVert \boldsymbol{p_i}(t) - \boldsymbol{p^d_i} \rVert^2$ and it represents the attraction of agent $i$ towards its goal position, with $\gamma_i^{-1}(0)$ being the desired set. The term $G_i(\boldsymbol{p}):\mathbb{R}^{3N}\rightarrow\mathbb{R}$ is associated with the collision avoidance property of agent $i$ with the rest of the team and is based on the inter-agent distance function \cite{Dimos_2007_journal}: $\beta_{ij}:\mathbb{R}^3\times\mathbb{R}^3\rightarrow\mathbb{R}$ with 
\begin{equation}
\beta_{ij}(\boldsymbol{p_i},\boldsymbol{p_j}) = \left\{ \begin{array}{ll} \lVert \boldsymbol{p_i}-\boldsymbol{p_j} \rVert^2-(r_i+r_j)^2, & \text{if } j\in\mathcal{N}_i\\ 
d^2_{s_i} -(r_i+r_j)^2, & \text{if } j\notin\mathcal{N}_i, \end{array} \right. \nonumber
\end{equation}
that represents the distance between agents $i$ and $j\in\mathcal{N}_i$. Roughly speaking, $G_i$ expresses all possible collisions between agent $i$ and the others and $G_i^{-1}(0)$ is the set we want to avoid. The term $f_i:\mathbb{R}\rightarrow\mathbb{R}$ is introduced in \cite{Dimos_2007_journal} and is used in this work in order to avoid inter-agent collisions in the case where $\gamma_{i}\rightarrow 0, \gamma_{j}\rightarrow 0, i,j\in\{1,\dots,N\}$ with $\boldsymbol{p^d_i}=\boldsymbol{p^d_j},i\neq j$, i.e., when two or more agents have the same goal positions and they approach them simultaneously. Analytic expressions for $G_i$ and $f_i$ can be found in \cite{Dimos_2007_journal}. With the aforementioned tools, the control law for agent $i$ is $\boldsymbol{u_i} = -k_i\dfrac{\partial \phi_i(\boldsymbol{p})}{\partial \boldsymbol{p_i}}$, which, as shown in \cite{Dimos_2007_journal}, drives all agents to their goal positions and guarantees inter-agent collision-avoidance.

\subsubsection{\textbf{Continuous Control Law}}
\mbox{}\\
By employing the aforementioned ideas regarding DNFs and given that $\mathcal{A}_i(\boldsymbol{p_i}(t_0))$ for some $t_0 \geq 0$, we propose a decentralized control law $\boldsymbol{u_i}$ for the transition $\pi_k\rightarrow_i\pi_{k'}$, as defined in Def. \ref{def:agent_transition}.
 
Initially, we define the set of "undesired" regions as $\Pi_{k,k'} = \{\pi_m\in\Pi, m\in\{1,\dots,K\} \backslash \{k ,k'\} \}$ and the corresponding free space $\mathcal{F}_{k,k'} = \mathcal{W}\backslash\{\mathcal{B}_{r_{\pi}}(\boldsymbol{p_{\pi}})\}_{\pi\in\Pi_{k,k'}}$. As the goal configuration we consider the centroid $\boldsymbol{p_{\pi_{k'}}}$ of $\pi_{k'}$ and we construct the function $\gamma_{i_{k'}}:\mathcal{F}_{k,k'}\rightarrow\mathbb{R}_{\geq 0}$ with $\gamma_{i_{k'}}(\boldsymbol{p_i}) = \lVert \boldsymbol{p_i} - \boldsymbol{p_{\pi_{k'}}} \rVert^2$. For the collision avoidance between the agents, we employ the function $G_i:\mathcal{F}_{k,k'}\times\mathbb{R}^{3(N-1)}\rightarrow\mathbb{R}$ as defined in \cite{Dimos_2007_journal}.

Moreover, we also need some extra terms that guarantee that agent $i$ will avoid the rest of the regions as well as the workspace boundary. To this end, we construct the function $\alpha_{i_{k,k'}}:\mathcal{F}_{k,k'}\rightarrow\mathbb{R}$ with $\alpha_{i_{k,k'}}(\boldsymbol{p_i}) = \alpha_{i,0}(\boldsymbol{p_i})\prod_{ m\in{\Pi}_{k,k'}}\alpha_{i,m}(\boldsymbol{p_i})$, where the function $\alpha_{i,0}:\mathcal{F}_{k,k'}\rightarrow\mathbb{R}$ is a measure of the distance of agent $i$ from the workspace boundary $\alpha_{i,0} = (r_0 - r_i)^2 - \lVert \boldsymbol{p_i} - \boldsymbol{p_0} \rVert^2$ and the function $\alpha_{i,m}:\mathcal{F}_{k,k'}\rightarrow\mathbb{R}$ is a measure of the distance of agent $i$ from the undesired regions $\alpha_{i,m} = \lVert \boldsymbol{p_i} - \boldsymbol{p_m} \rVert^2 - (r_i + r_m)^2$. 

With the above ingredients, we construct the following navigation function $\phi_{i_{k,k'}}:\mathcal{F}_{k,k'}\times \mathbb{R}^{3(N-1)}\rightarrow[0,1]$:
\begin{equation}
\phi_{i_{k,k'}}(\boldsymbol{p}(t)) = \dfrac{\gamma_{i_{k'}}(\boldsymbol{p_i}) + f_i(G_i)}{ ( \gamma_{i_{k'}}^{\lambda_i}(\boldsymbol{p_i}) + G_i(\boldsymbol{p})\alpha_{i_{k,k'}}(\boldsymbol{p_i}) )^{1/\lambda_i}} \label{eq:navigation function}
\end{equation}
for agent $i$, with $\lambda_i > 0$ and the following vector field: 
\begin{equation}
\boldsymbol{c_{i_{k,k'}}}(t) = \left\{ \begin{array}{cc}  -k_{g_i}\dfrac{\partial \phi_{i_{k,k'}}(\boldsymbol{p}(t))}{\partial\boldsymbol{p_i}(t)}, & \mbox{if }  \pi_{k} \not \equiv  \pi_{k'} \\  
0 &  \mbox{if } \pi_{k} \equiv  \pi_{k'} 
\end{array} \right. \label{eq:feedback_contr} 
\end{equation}
for all $t \geq t_0$, with $k_{g_i}>0$ and $f_i(G_i)$ as defined in \cite{Dimos_2007_journal}.

The navigation field (\ref{eq:feedback_contr}) guarantees that agent $i$ will not enter the undesired regions or collide with the other agents and $\lim_{t\rightarrow\infty}\boldsymbol{p_i}(t) = \boldsymbol{p_{\pi_{k'}}}$. The latter property of asymptotic convergence along with the assumption that $r_i < r_{\pi_k}, \forall i\in\{1,\dots,N\},k\in\{1,\dots,K\}$, implies that there exists a finite time instant $t^{\scriptscriptstyle f}_{i,k'} \geq t_0$ such that $\boldsymbol{p_i}(t^{\scriptscriptstyle f}_{i,k'}) \in\mathcal{B}_{r_{\pi_{k'}}}(\boldsymbol{p_{\pi_{k'}}})$ and more specifically that  $\mathcal{A}_i(\boldsymbol{p_i}(t^{\scriptscriptstyle f}_{i,k'}))\in\pi_{k'}$, which is the desired behavior. 
The time instant $t^{\scriptscriptstyle f}_{i,k'}$ can be chosen from the set $S_{t_{k'}} = \{t\geq t_0, \mathcal{A}_i(\boldsymbol{p_i}(t))\in\pi_{k'}\}$. 

Note, however, that once agent $i$ leaves region $\pi_k$, there is no guarantee that it will not enter that region again (note that $F_{k,k'}$ includes $\pi_k$). Therefore, we define the set $\Pi_{\emptyset,k'} = \{\pi_m\in\Pi, m\in\{1,\dots,K\} \backslash \{k'\}\}$ and the corresponding free space $\mathcal{F}_{\emptyset,k'} = \mathcal{W}\backslash\{\mathcal{B}_{r_{\pi}}(\boldsymbol{p_{\pi}})\}_{\pi\in\Pi_{\emptyset,k'}}$, and we construct the function $\phi_{i_{\emptyset,k'}}:\mathcal{F}_{\emptyset,k'}\times \mathbb{R}^{3(N-1)}\rightarrow[0,1]$:
\begin{equation}
\phi_{i_{\emptyset,k'}}(\boldsymbol{p}(t)) = \dfrac{\gamma_{i_{k'}}(\boldsymbol{p_i})+f_i(G_i)}{ ( \gamma_{i_{k'}}^{\lambda_i}(\boldsymbol{p_i}) + G_i(\boldsymbol{p})\alpha_{i_{\emptyset,k'}}(\boldsymbol{p_i}) )^{1/\lambda_i}} \label{eq:navigation function_2}
\end{equation}
where $\alpha_{i_{\emptyset,k'}}=\alpha_{i,0}(\boldsymbol{p_i})\prod_{ m\in\Pi_{\emptyset,k'}}\alpha_{i,m}(\boldsymbol{p_i})$, with corresponding vector field: 
 \begin{equation}
 \boldsymbol{c_{i_{\emptyset,k'}}}(t) =  -k_{g_i}\dfrac{\partial \phi_{i_{\emptyset,k'}}(\boldsymbol{p}(t))}{\partial\boldsymbol{p_i}(t)}, \label{eq:feedback_contr_2}
 \end{equation}
 which guarantees that region $\pi_k$ will be also avoided. Therefore, we develop a switching control protocol that employs (\ref{eq:feedback_contr}) until agent $i$ is out of region $\pi_k$ and then switches to (\ref{eq:feedback_contr_2}) until $t=t^{\scriptscriptstyle f}_{i,k'}$. Consider the following switching function:
 \begin{equation}
 s(x) = \dfrac{1}{2}(\text{sat}(2x-1)+1)
 \end{equation} 
 where $\text{sat}:\mathbb{R}\rightarrow[-1,1]$ is the standard saturation function ($\text{sat}(x)=x$, if $\lvert x \rvert \leq 1; \text{sat}(x)=x/\lvert x \rvert$, if $\lvert x \rvert > 1$),  
 and the time instant $t'_{i,k}$ that represents the moment that agent $i$ is out of region $\pi_k$, i.e., $t'_{i,k} = \min S_{t_{\not k}}$, where $S_{t_{\not k}} = \{t\geq t_0,  \mathcal{B}_{r_i}(\boldsymbol{p_i}(t))\cap\mathcal{B}_{r_{\pi_k}}(\boldsymbol{p_{\pi_k}})=\emptyset \}$. Note that $t'_{i,k} < t^{\scriptscriptstyle f}_{i,k'}$, since $d_3(\boldsymbol{p_{\pi_k}},\boldsymbol{p_{\pi_{k'}}}) > 4\max_{k\in\{1,\dots,K\}}(r_{\pi_k}), \forall k,k'\in\{1,\dots,K\}$ with $k\neq k'$.
 Then, we propose the following switching control protocol $\boldsymbol{u_i}:[t_0,t^{\scriptscriptstyle f}_{i,k'})\rightarrow \mathbb{R}^3$:
 \small
 \begin{equation}
 \boldsymbol{u_i}(t) = \left\{ \begin{array}{cc} \boldsymbol{c_{i_{k,k'}}}(t), &  t\in T_1 \\
							(1-s(\xi_{i,k}))\boldsymbol{c_{i_{k,k'}}}(t) + s(\xi_{i,k})\boldsymbol{c_{i_{\emptyset,k'}}}(t), &  t\in T_2 
								\end{array} \right.
 \label{eq:switch_controller}
 \end{equation}
 \normalsize
 where $T_1 = [t_0,t'_{i,k}), T_2 = [t'_{i,k},t^{\scriptscriptstyle f}_{i,k'})$ and $\xi_{i,k} =$ \small$\dfrac{t-t'_{i,k}}{\nu_i}$\normalsize, where $\nu_i$ is a design parameter indicating the time period of the switching process, with $t^{\scriptscriptstyle f}_{i,k'}-t'_{i,k}>\nu_i > 0$. Invoking the continuity of $\boldsymbol{p_i}(t)$, we obtain $\lim_{t\rightarrow (t^{\scriptscriptstyle f}_{i,k'})^-}\boldsymbol{p_i}(t) = \boldsymbol{p_i}(t^{\scriptscriptstyle f}_{i,k'})\in\mathcal{B}_{r_{\pi_{k'}}}(\boldsymbol{p_{\pi_{k'}}})$ and hence the control protocol (\ref{eq:switch_controller}) guarantees, for sufficiently small $\nu_i$, that agent $i$ will navigate from $\pi_k$ to $\pi_{k'}$ in finite time without entering any other regions or colliding with other agents and therefore establishes a transition $\pi_k\rightarrow_i\pi_{k'}$. The proof of correctness of (\ref{eq:navigation function}) and (\ref{eq:navigation function_2}) follows closely the one in \cite{Dimos_2007_journal} and is therefore omitted.
 
\subsection{High-Level Plan Generation} \label{subsec:High level plan}

The next step of our solution is the high-level plan, which can be generated using standard techniques inspired by automata-based formal verification methodologies. In Section \ref{subsec:Continuous Control}, we proposed a continuous control law that allows the agents to transit between any $\pi_k, \pi_{k'}\in\Pi$ in the given workspace $\mathcal{W}$, without colliding with each other. Thanks to this and to our definition of LTL semantics over the sequence of atomic propositions, we can abstract the motion capabilities of each agent as a finite transition system $\mathcal{T}_i$ as follows \cite{BayeKatoen_2008}:
\begin{defn}
The motion of each agent $i\in\{1,\dots,N\}$ in $\mathcal{W}$ is modeled by the following Transition System (TS):
\begin{equation}
\mathcal{T}_i=(\Pi_i,\Pi^{\text{init}}_i, \rightarrow_i, \Psi_i, \mathcal{L}_i),\label{eq:TS_i}
\end{equation}
where $\Pi_i\subseteq\Pi$ is the set of states represented by the regions of interest that the agent can be at, according to Def. \ref{def:agent in region}, $\Pi^{\text{init}}_i \subseteq \Pi_i $ is the set of initial states that agent $i$ can start from, $\rightarrow_i\subseteq\Pi_i\times\Pi_i$ is the transition relation established in Section \ref{subsec:Continuous Control}, abbreviated as $\pi_k\rightarrow\pi_{k'}, \pi_k,\pi_{k'}\in\Pi_i$, and $\Psi_i, \mathcal{L}_i$ are the atomic propositions and labeling function respectively, as defined in Section \ref{sec:System and PF}.
\end{defn}

After the definition of $\mathcal{T}_i$, we translate each given LTL formula $\phi_i, i\in\{1,\dots,N\}$ into a Büchi automaton $\mathcal{C}_i$ and we form the product $\widetilde{\mathcal{T}}_i=\mathcal{T}_i\times\mathcal{C}_i$. The accepting runs of $\widetilde{\mathcal{T}}_i$ satisfy $\phi_i$ and are directly projected to a sequence of waypoints to be visited, providing therefore a desired path for agent $i$. Although the semantics of LTL is defined over infinite sequences of atomic propositions, it can be proven that there always exists a high-level plan that takes a form of a finite state sequence followed by an infinite repetition of another finite state sequence. For more details on the followed technique, we kindly refer the reader to the related literature, e.g., \cite{BayeKatoen_2008}.

Following the aforementioned methodology, we obtain a high-level plan for each agent as sequences of regions and atomic propositions $p_i = \pi_{i_1} \pi_{i_2} \dots$ and $\psi_i = \psi_{i_1} \psi_{i_2}\dots$ with $i_m\in\{1,\dots,K\}, \psi_{i_m}\in2^{\Psi_i}, \psi_{i_m}\in\mathcal{L}_i(\pi_{i_m}), \forall m\in\mathbb{N}$ and $\psi_i\models\phi_i,\forall i\in\{1,\dots,N\}$. 

The execution of $(p_i,\psi_i)$ produces a trajectory $\boldsymbol{p_i}(t)$ that corresponds to the behavior $\beta_i = (\boldsymbol{p_i}(t),\psi_i) = (\boldsymbol{p_{i_1}}(t),\psi_{i_1})(\boldsymbol{p_{i_2}}(t),\psi_{i_2})\dots$, with $\mathcal{A}_i(\boldsymbol{p_{i_m}})\in\pi_{i_m}$ and $\psi_{i_m}\in\mathcal{L}_i(\pi_{i_m})$, $\forall m\in\mathbb{N}$. Therefore, since $\psi_i\models \phi_i$, the behavior $\beta_i$ yields satisfaction of the formula $\phi_i$. Moreover, the property of inter-agent collision avoidance is inherent in the transition relations of $\mathcal{T}_i$ and guaranteed by the navigation control algorithm of Section \ref{subsec:Continuous Control}. 
The previous discussion is summarized in the following theorem:
\begin{thm}
The individual executions of $(p_i,\psi_i), i\in\{1,\dots,N\}$, that satisfy the respective $\phi_i$, produce agent behaviors $\beta_i, i\in\{1,\dots,N\}$ that (i) yield the satisfaction of all $\phi_i, i\in\{1,\dots,N\}$ and (ii) guarantee inter-agent collision avoidance, providing, therefore, a solution to Problem \ref{problem}. 
\end{thm}

 \begin{rem}
The proposed control algorithm is decentralized in the sense that each agent derives and executes its own plan without communicating with the rest of the team. The only  information that each agent has is the position of its neighboring agents that lie in its limited sensing radius. 
 \end{rem}

\begin{figure}[!btp]
\centering
\includegraphics[trim = 0cm 0.5cm 0cm -1cm,width = 0.33\textwidth, height = 0.23\textwidth]{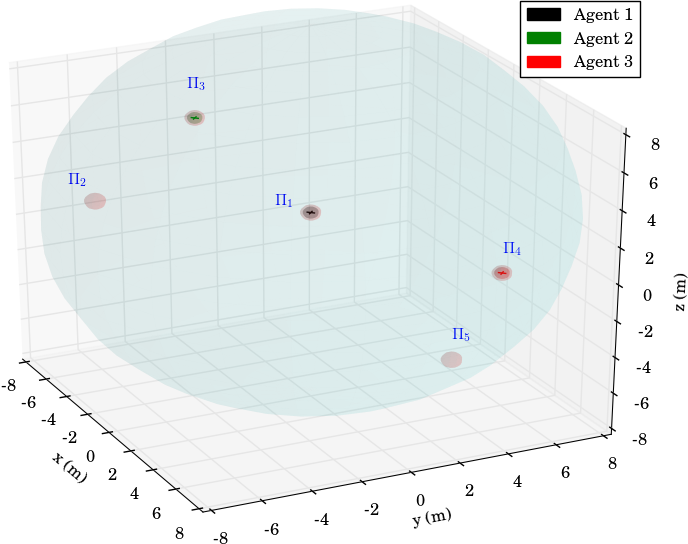}

\caption{Initial workspace of the simulation studies. The grey spheres represent the regions of interest while the black, green and red crosses represent agents 1,2 and 3, respectively, along with their bounding spheres. \label{fig:Initial_workspace}}
\end{figure}

\begin{figure}[!btp]
\centering
\includegraphics[trim = 0cm 0cm 0cm 0cm,width = 0.35\textwidth, width = 0.35\textwidth]{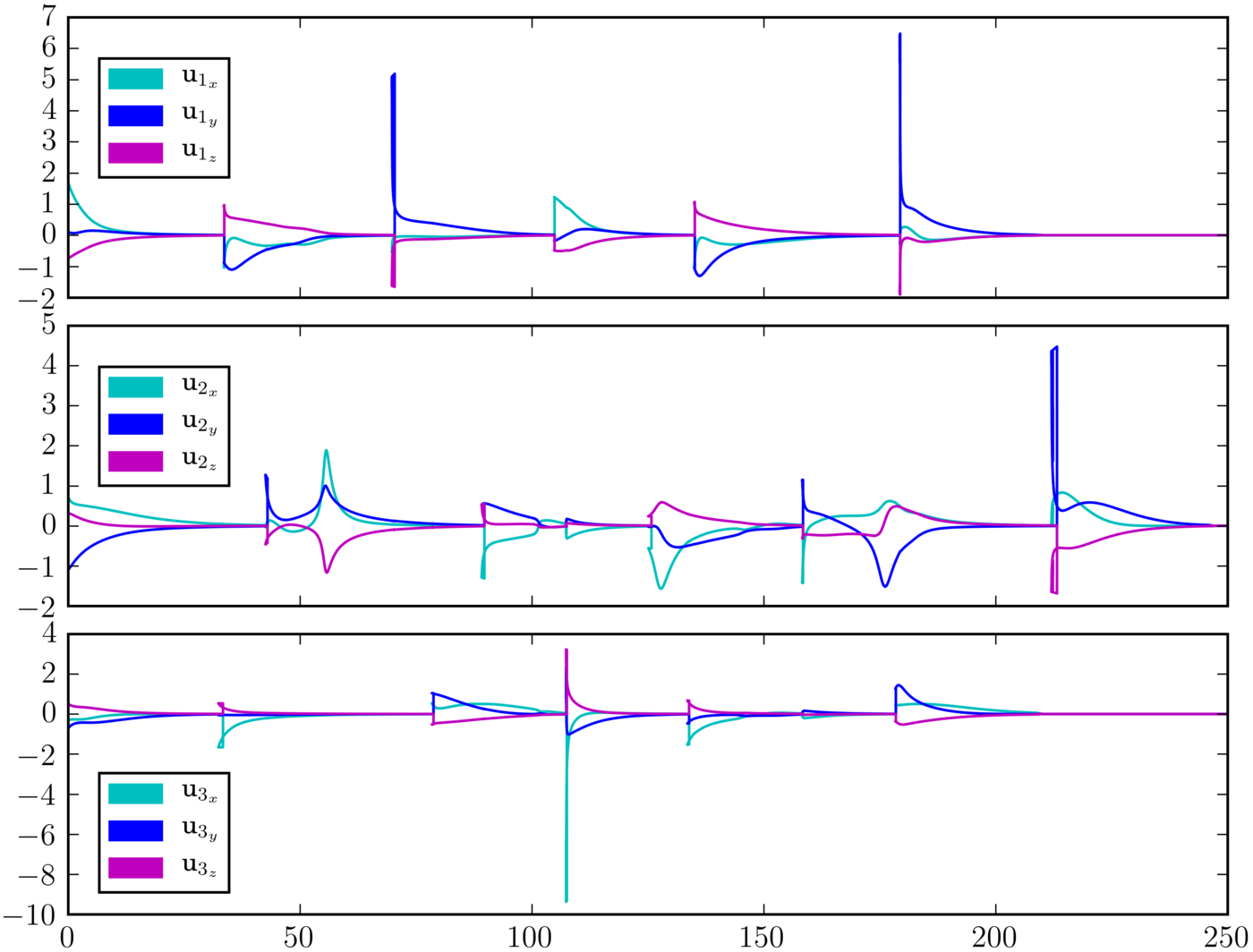}

\caption{The resulting $3$-dimensional control signals of the $3$ agents for the simulation studies. Top: agent $1$, middle: agent $2$, bottom: agent $3$.\label{fig:sim_vel}}
\end{figure}

\section{Simulation Results} \label{sec:simulation results}
To demonstrate the efficiency of the proposed algorithm, we consider $N=3$ aerial vehicles $\mathcal{B}_{r_i}(\boldsymbol{p_i}(t))$, with $r_i = 0.3$m, $d_{s_i} = 0.65$m,  $\forall i=\{1,2,3\}$, operating in a workspace $\mathcal{W} = \mathcal{B}_{r_0}(\boldsymbol{p_0})$ with $r_0 = 10$m and $\boldsymbol{p_0} = [0,0,0]^T$m. Moreover, we consider $K=5$ spherical regions of interest $\mathcal{B}_{r_{\pi_k}}(\boldsymbol{p_{\pi_k}})$ with $r_{\pi_k} = 0.4$m, $\forall k=\{1,\dots,5\}$ and $\boldsymbol{p}_{\pi_1} = [0,0,2]^T$m, $\boldsymbol{p}_{\pi_2} = [1,-9,5]^T$m, $\boldsymbol{p}_{\pi_3} = [-8,-1,4]^T$m, $\boldsymbol{p}_{\pi_4} = [2,7,-2]^T$m and $\boldsymbol{p}_{\pi_5} = [7.5,2,-3]^T$m.  The initial configurations of the agents are taken as $\boldsymbol{p_1}(0) = \boldsymbol{p}_{\pi_1}, \boldsymbol{p_2}(0) = \boldsymbol{p}_{\pi_3}, \boldsymbol{p_3}(0) = \boldsymbol{p}_{\pi_4}$ and therefore, $\mathcal{A}_1(\boldsymbol{p_1}(0))\in\pi_1, \mathcal{A}_2(\boldsymbol{p_2}(0))\in\pi_3$ and  $\mathcal{A}_3(\boldsymbol{p_3}(0))\in\pi_4$. An illustration of the described workspace is depicted in Fig. \ref{fig:Initial_workspace}.

\begin{figure}[!btp]
\centering
  \subfloat[]{\includegraphics[trim = 0cm -0.2cm 0cm -1cm,width=0.2\textwidth,height=0.1475\textwidth]{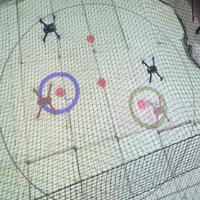}\label{fig:Initial_workspace_exp_SML}}
  \quad
  \subfloat[]{\includegraphics[trim = 0cm 1cm 0cm -1cm,width=0.25\textwidth, height=0.155\textwidth]{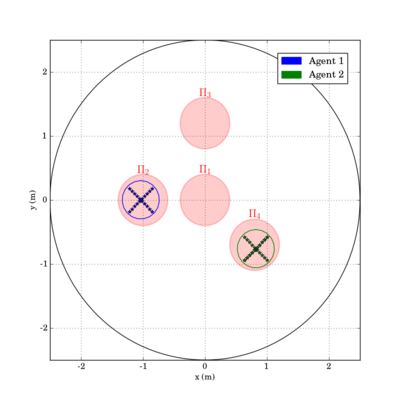}\label{Initial_workspace_exp_simulated}}
  \caption{Initial workspace for the first real experimental scenario. (a): The UAVs with the projection of their bounding spheres, (with blue and green), and the centroids of the regions of interest (with red). (b): Top view of the described workspace. The UAVs are represented by the blue and green circled X's and the regions of interest by the red disks $\pi_1,\dots,\pi_4$.}\label{fig:Initial_workspace_exp_1}
\end{figure}

\begin{figure*}[!tbp]
  \centering
  \subfloat[]{\includegraphics[trim = 0cm 0.5cm 0cm -1cm,width=0.23\textwidth]{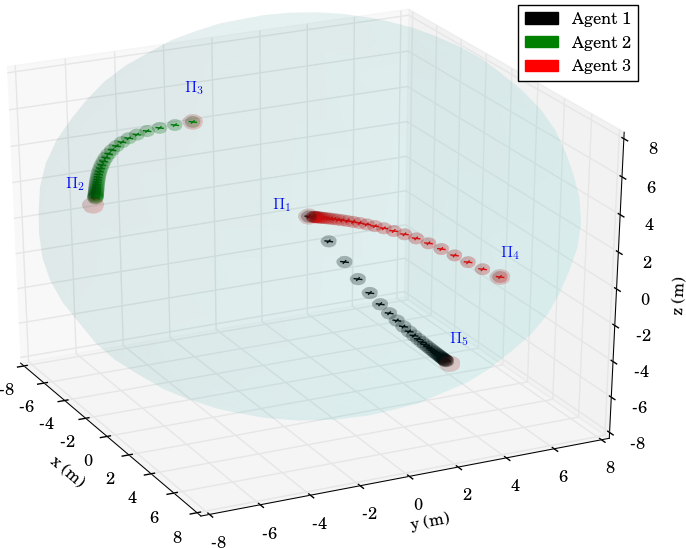}\label{fig:path:1}}
  \quad
  \subfloat[]{\includegraphics[trim = 0cm 0.5cm 0cm -1cm,width=0.23\textwidth]{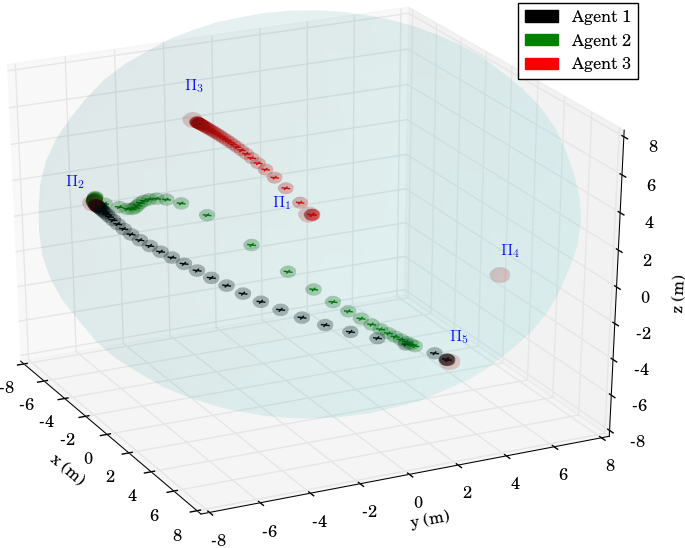}\label{fig:path:2}}
  \quad
  \subfloat[]{\includegraphics[trim = 0cm 0.5cm 0cm -1cm,width=0.23\textwidth]{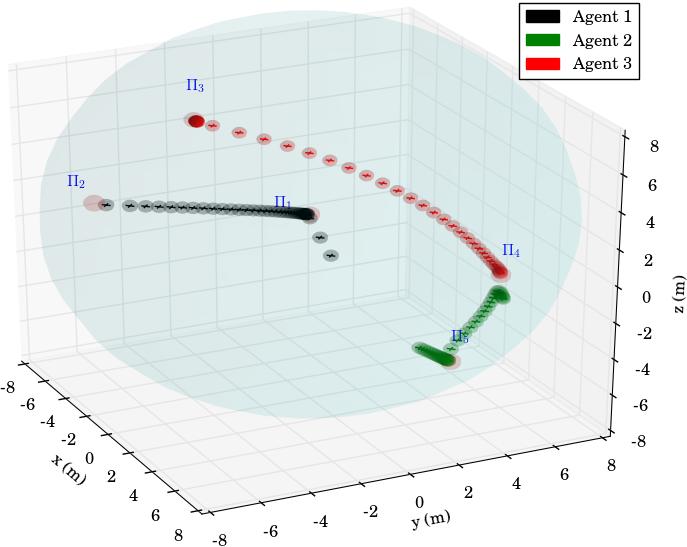}\label{fig:path:3}}
  \quad
  \subfloat[]{\includegraphics[trim = 0cm 0.5cm 0cm -1cm,width=0.23\textwidth]{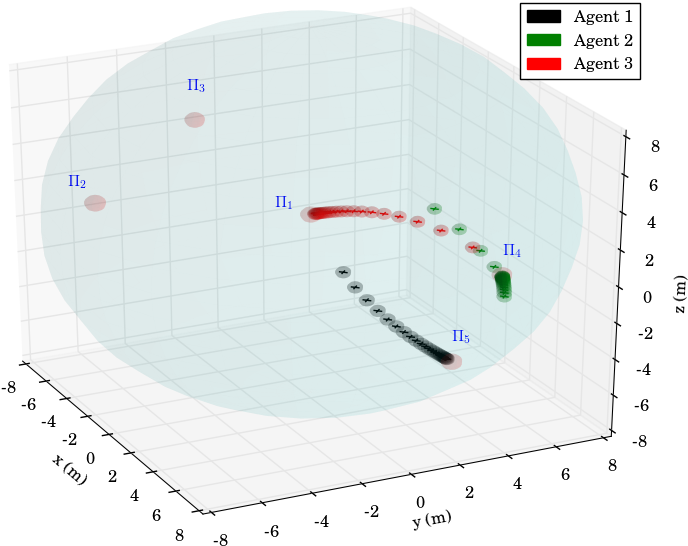}\label{fig:path:4}}
  \hfill
  \subfloat[]{\includegraphics[trim = 0cm 0cm 0cm 0cm,width=0.23\textwidth]{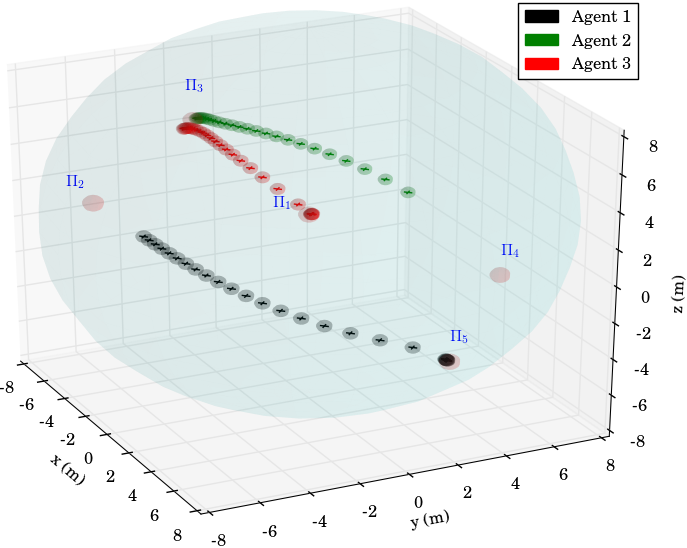}\label{fig:path:5}}
  \quad
  \subfloat[]{\includegraphics[trim = 0cm 0cm 0cm 0cm,width=0.23\textwidth]{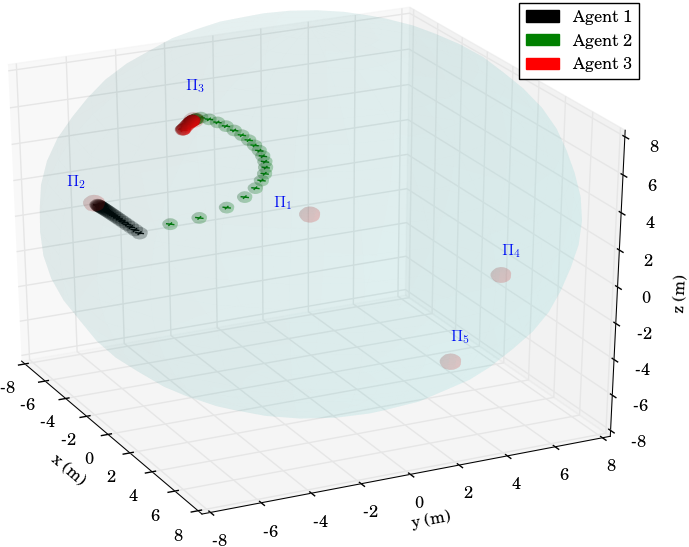}\label{fig:path:6}}
  \quad
  \subfloat[]{\includegraphics[trim = 0cm 0cm 0cm 0cm,width=0.23\textwidth]{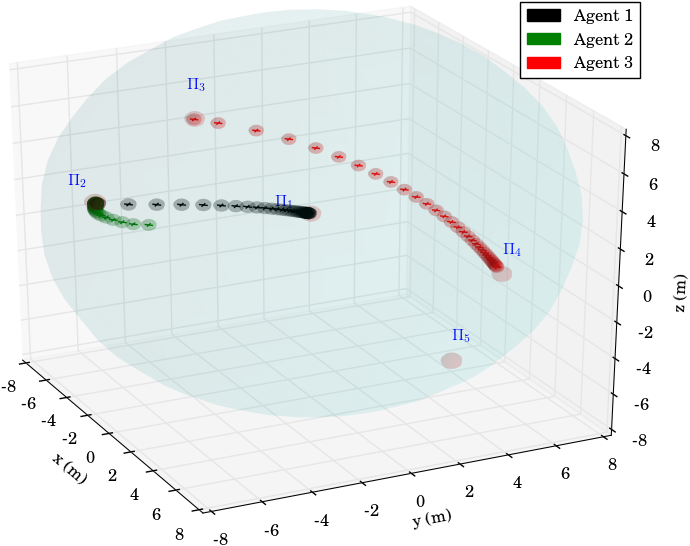}\label{fig:path:7}}
  \quad
  \subfloat[]{\includegraphics[trim = 0cm 0cm 0cm 0cm,width=0.23\textwidth]{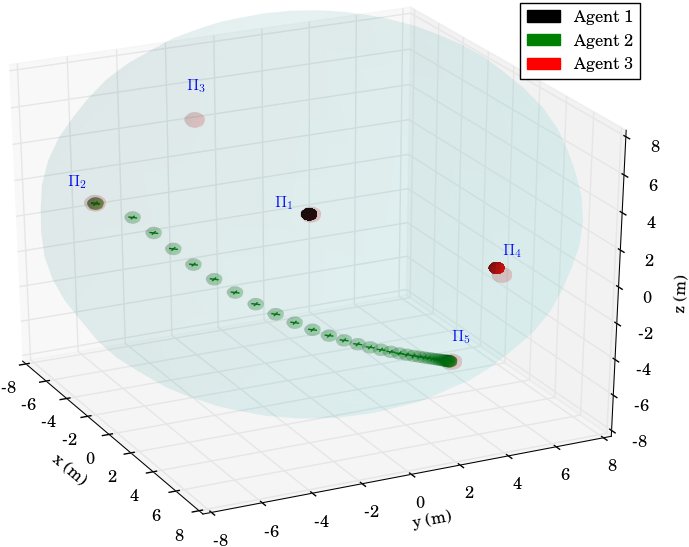}\label{fig:path:8}}
  \caption{Execution of the paths $(\pi_1\pi_5\pi_2)^2\pi_1, (\pi_3\pi_2\pi_5\pi_4)^2\pi_3\pi_2\pi_5$ and $(\pi_4\pi_1\pi_3)^2\pi_4$ by agents $1,2$ and $3$, respectively, for the simulation studies. }\label{fig:path}
\end{figure*}

 We consider that agent $2$ is assigned with inspection tasks and has the  atomic propositions $\Psi_2 = \{ ``\text{ins}_{\text{a}}",``\text{ins}_{\text{b}}",``\text{ins}_{\text{c}}",``\text{ins}_{\text{d}}",``\text{obs}" \}$ with $\mathcal{L}_{2}(\pi_1) = \{``\text{obs}"\}, \mathcal{L}_{2}(\pi_2) = \{``\text{ins}_{\text{a}}"\}, \mathcal{L}_{2}(\pi_3) = \{``\text{ins}_{\text{b}}"\}, \mathcal{L}_{2}(\pi_4) = \{``\text{ins}_{\text{c}}"\}$ and $\mathcal{L}_{2}(\pi_5) = \{``\text{ins}_{\text{d}}"\}$, where we have considered that region $\pi_1$ is an undesired ("obstacle") region for this agent. More specifically, the task for agent $2$ is the continuous inspection of the workspace while avoiding region $\pi_1$. The corresponding LTL specification is $\phi_2 =  (\square \neg ``\text{obs}")\land\square(\lozenge``\text{ins}_{\text{a}}"\land\lozenge``\text{ins}_{\text{b}}"\land\lozenge``\text{ins}_{\text{c}}"\land\lozenge``\text{ins}_{\text{d}}")$. 
 Agents $1$ and $3$ are interested in moving around resources scattered in the workspace and have propositions $\Psi_1 = \Psi_3 = \{ ``\text{res}_{\text{a}}",``\text{res}_{\text{b}}",``\text{res}_{\text{c}}",``\text{res}_{\text{d}}", ``\text{res}_{\text{e}}" \}$ with $\mathcal{L}_1(\pi_1)=\mathcal{L}_3(\pi_1)=\{\text{res}_{\text{a}}\}, \mathcal{L}_1(\pi_2)=\mathcal{L}_3(\pi_2)=\{\text{res}_{\text{b}}\}, \mathcal{L}_1(\pi_3)=\mathcal{L}_3(\pi_3)=\{\text{res}_{\text{c}}\}, \mathcal{L}_1(\pi_4)=\mathcal{L}_3(\pi_4)=\{\text{res}_{\text{d}}\}$ and $\mathcal{L}_1(\pi_5)=\mathcal{L}_3(\pi_5)=\{\text{res}_{\text{e}}\}$. We assume that $``\text{res}_{\text{a}}"$ is shared between the two agents whereas $``\text{res}_{\text{b}}"$ and $``\text{res}_{\text{e}}"$ have to be accessed only by agent $1$ and $``\text{res}_{\text{c}}"$ and $``\text{res}_{\text{d}}"$ only by agent $3$. The corresponding specifications are $\phi_1 = \square \neg (``\text{res}_{\text{c}}"\lor``\text{res}_{\text{d}}")\land \square\lozenge(``\text{res}_{\text{a}}" \bigcirc``\text{res}_{\text{e}}"\bigcirc``\text{res}_{\text{b}}")$ and $\phi_3 = \square \neg (``\text{res}_{\text{b}}"\lor``\text{res}_{\text{e}}")\land 	\square\lozenge(``\text{res}_{\text{a}}" \bigcirc``\text{res}_{\text{c}}"\bigcirc``\text{res}_{\text{d}}")$, where we have also included a specific order for the access of the resources. Next, we employ the off-the-shelf tool LTL2BA \cite{LTL2BA} to create the Büchi automata $\mathcal{C}_i, i=\{1,2,3\}$ and by following the procedure described in Section \ref{subsec:High level plan}, we derive the paths $p_1 = (\pi_1\pi_5\pi_2)^{\omega}, p_2 = (\pi_3\pi_2\pi_5\pi_4)^{\omega}, p_3 = (\pi_4\pi_1\pi_3)^{\omega}$, whose execution satisfies $\phi_1,\phi_2,\phi_3$. Regarding the continuous control protocol, we chose $k_{g_i} = 15, \lambda_i = 5, \forall i\in\{1,2,3\}$ in (\ref{eq:feedback_contr}), (\ref{eq:feedback_contr_2}) and the switching duration in (\ref{eq:switch_controller}) was calculated online as $\nu_i = 0.1t'_{i,k}$, where we assume that the large distance between the regions $\pi_k$ (see Fig. \ref{fig:Initial_workspace}) implies that $t^{\scriptscriptstyle f}_{i,k'} > 1.1t'_{i,k}$ and thus, $\nu_i < t^{\scriptscriptstyle f}_{i,k'} - t'_{i,k}$. The simulation results are depicted in Fig. \ref{fig:sim_vel} and \ref{fig:path}. In particular, Fig. \ref{fig:path} illustrates the execution of the paths $(\pi_1\pi_5\pi_2)^2\pi_1, (\pi_3\pi_2\pi_5\pi_4)^2\pi_3\pi_2\pi_5$ and $(\pi_4\pi_1\pi_3)^2\pi_4$ by agents $1,2$ and $3$ respectively, where the superscript $2$ here denotes that the corresponding paths are executed twice. Fig. \ref{fig:sim_vel} depicts the resulting control inputs $u_i, \forall i\in\{1,2,3\}$. The figures demonstrate the successful execution of the agents' paths and therefore, satisfaction of the respective formulas with inter-agent collision avoidance.

\begin{figure*}[!tbp]
  \centering
  \subfloat[]{\includegraphics[trim = 0cm 0cm 0cm 0cm,width=0.25\textwidth,height = 0.2\textwidth]{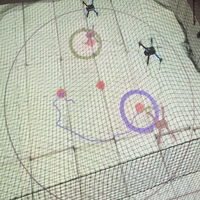}\label{fig:exp_1_path:1}}
  \quad
  \subfloat[]{\includegraphics[trim = 0cm 0cm 0cm 0cm,width=0.25\textwidth,height = 0.2\textwidth]{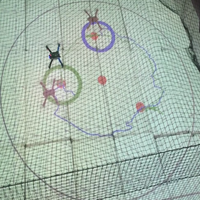}\label{fig:exp_1_path:2}}
  \quad
  \subfloat[]{\includegraphics[trim = 0cm 0cm 0cm 0cm,width=0.25\textwidth,height = 0.2\textwidth]{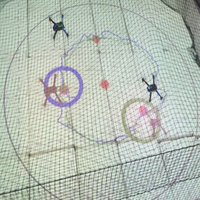}\label{fig:exp_1_path:3}}
  \hfill
  \subfloat[]{\includegraphics[trim = 1cm 0cm 0cm 0cm,width=0.25\textwidth, height = 0.2\textwidth]{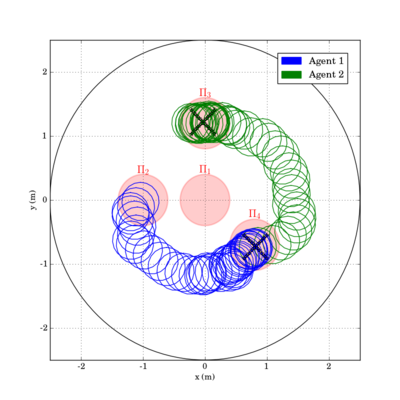}\label{fig:exp_1_path:1_sim}}
  \quad
  \subfloat[]{\includegraphics[trim = 1cm 0cm 0cm 0cm,width=0.25\textwidth, height = 0.2\textwidth]{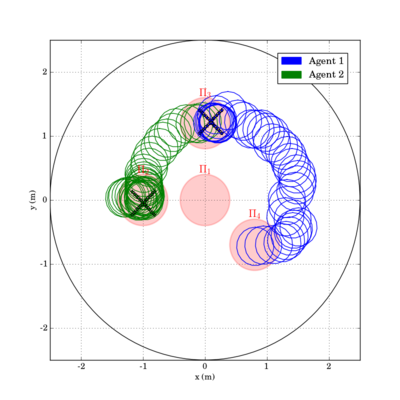}\label{fig:exp_1_path:2_sim}}
  \quad
  \subfloat[]{\includegraphics[trim = 1cm 0cm 0cm 0cm,width=0.25\textwidth, height = 0.2\textwidth]{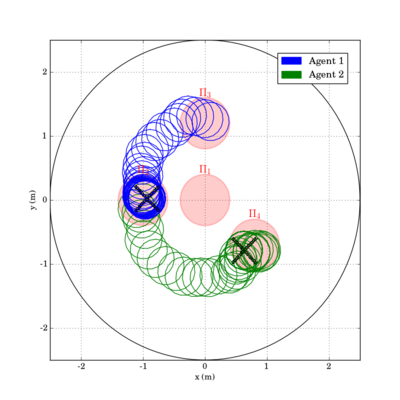}\label{fig:exp_1_path:3_sim}}
  \caption{Execution of the paths $(\pi_2\pi_4\pi_3)^1$ and $(\pi_4\pi_3\pi_2)^1$ by agents $1$ and $2$, respectively for the first experimental scenario. (a), (d): $\pi_2\rightarrow_1\pi_4, \pi_4\rightarrow_2\pi_3$, (b), (e): $\pi_4\rightarrow_1\pi_3, \pi_3\rightarrow_2\pi_2$, (c), (f):$\pi_3\rightarrow_1\pi_2, \pi_2\rightarrow_2\pi_4$.  }\label{fig:exp_1_path}
\end{figure*}

\begin{figure}[!btp]
\centering
\includegraphics[trim = 0cm 0cm 0cm -0.5cm,width = 0.45\textwidth]{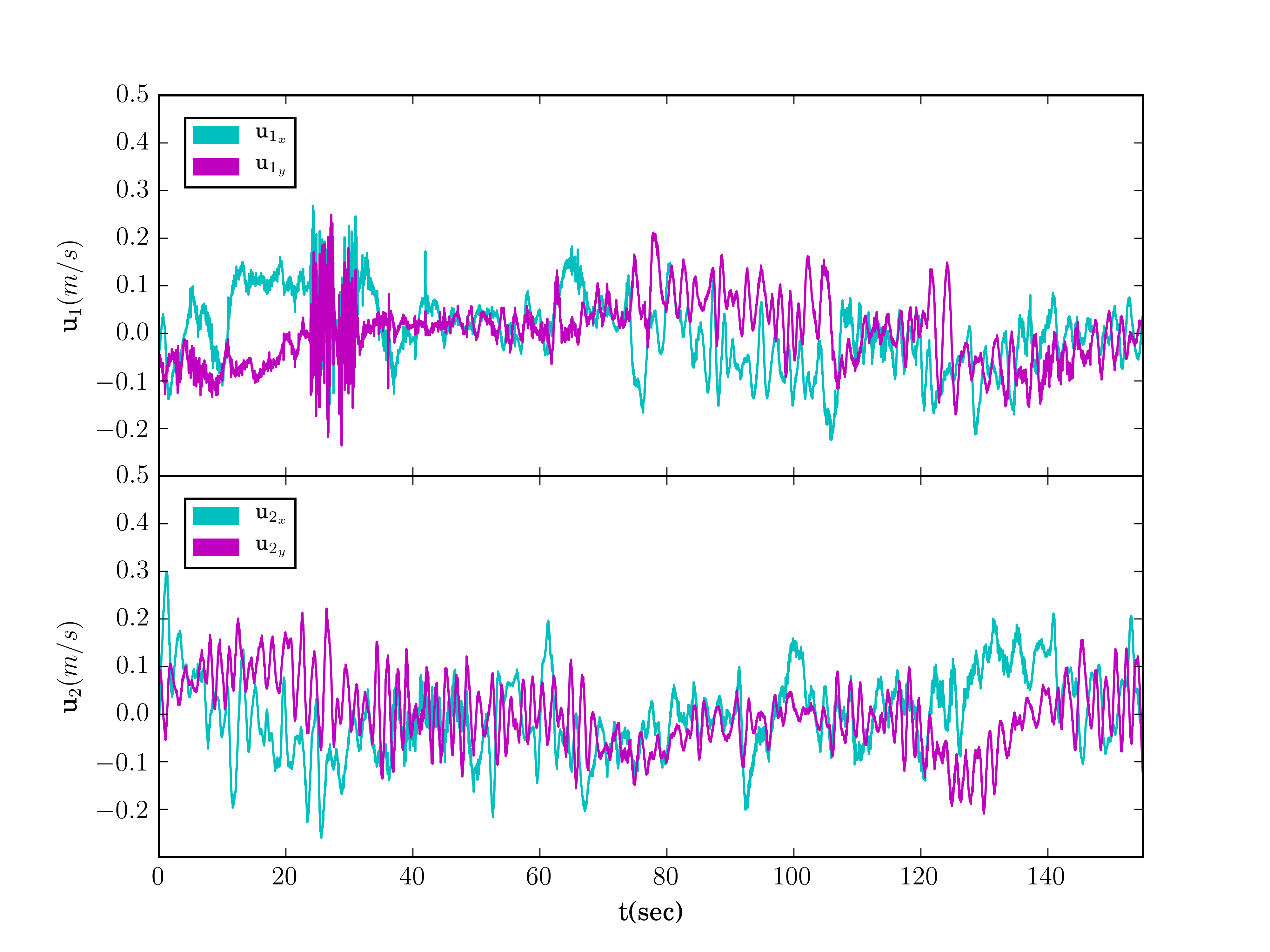}

\caption{The resulting $2$-dimensional control signals of the $2$ agents for the first experimental scenario. Top: agent 1, bottom: agent 2. \label{fig:exp_1_vel}}
\end{figure}

\section{Experimental Results} \label{sec:experimental results}

The validity and efficiency of the proposed solution was also verified through real-time experiments. The experimental setup involved two remotely controlled \textit{IRIS+} quadrotors from $3$D Robotics, which we consider to have sensing range $d_{s_i} = 0.65$m, upper control input bound $\lvert \boldsymbol{u}_m \rvert \leq 1$m/s, $m\in\{x,y,z\}$, and bounding spheres with radius $r_i = 0.3$m, $\forall i\in\{1,2\}$. We considered two $2$-dimensional scenarios in a workspace $\mathcal{W} = \{\boldsymbol{p}\in\mathbb{R}^2 \text{ s.t. } \lVert \boldsymbol{p} \rVert \leq 2.5\text{m} \}$, i.e. $\boldsymbol{p_0} = [0,0]^T$ and $r_0 = 2.5$m. 

The first scenario included $4$ regions of interest $\Pi=\{\pi_1,\dots,\pi_4\}$ in $\mathcal{W}$, with $r_{\pi_k} = 0.4,\forall k\in\{1,\dots,4\}$ and $\boldsymbol{p_{\pi_1}} = [0,0]^T$m, $\boldsymbol{p_{\pi_2}} = [-1,0]^T$m, $\boldsymbol{p_{\pi_3}} = [0,1.25]^T$m and $\boldsymbol{p_{\pi_4}} = [0.8,-0.7]^T$m. The initial positions of the agents were taken such that $\mathcal{A}_1(\boldsymbol{p_1}(0)) \in\pi_2$ and $\mathcal{A}_2(\boldsymbol{p_2}(0))\in\pi_4$ (see Fig. \ref{fig:Initial_workspace_exp_1}). We also defined the atomic propositions $\Psi_1 = \Psi_2 = \{``\text{obs}",``a",``b",``c"\}$ with $L_1(\pi_1) = L_2(\pi_1)=\{``\text{obs}"\}, L_1(\pi_2) = L_2(\pi_2) = \{``a"\}, L_1(\pi_3) = L_2(\pi_3)=\{``b"\}, L_1(\pi_4) = L_2(\pi_4)=\{``c"\}$. In this scenario, we were interested in area inspection while avoiding the "obstacle" region, and thus, we defined the individual specifications with the following LTL formulas: $\phi_1 = \phi_2 = \square\neg``\text{obs}"\land\square\lozenge(``a"\bigcirc``c"\bigcirc``b")$. By following the procedure described in Section \ref{subsec:High level plan}, we obtained the paths $p_1 = (\pi_2\pi_4\pi_3)^{\omega}, p_2 = (\pi_4\pi_2\pi_3)^{\omega}$. Fig. \ref{fig:exp_1_path} depicts the execution of the paths $(\pi_2\pi_4\pi_3)^1$ and $(\pi_4\pi_2\pi_3)^1$ by agents $1$ and $2$, respectively, and Fig. \ref{fig:exp_1_vel} shows the corresponding input signals, which do not exceed the control bounds $1$m/s. It can be deduced by the figures that the agents successfully satisfy their individual formulas, without colliding with each other. 

\begin{figure}[!btp]
\centering
  \subfloat[]{\includegraphics[trim = 0cm -0.2cm 0cm -1cm,width=0.2\textwidth,height=0.17\textwidth]{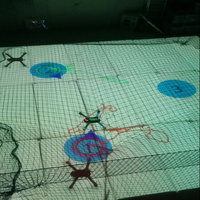}\label{fig:Initial_workspace_exp2_SML}}
  \quad
  \subfloat[]{\includegraphics[trim = 0cm 1cm 0cm -1cm,width=0.25\textwidth, height=0.18\textwidth]{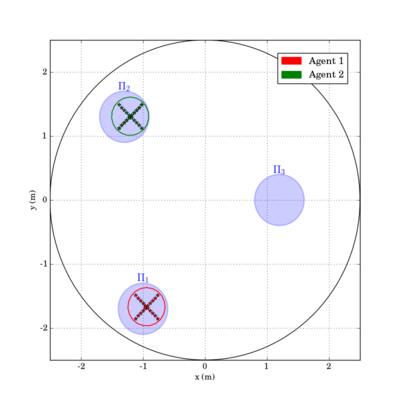}\label{Initial_workspace_exp2_simulated}}
  \caption{Initial workspace for the second experimental scenario. (a): The UAVs with the projection of their bounding spheres, (with red and green), and the regions of interest (blue disks). (b): Top view of the described workspace. The UAVs are represented by the red and green circled X's and the regions of interest by the blue disks $\pi_1,\dots,\pi_3$.}\label{fig:Initial_workspace_exp_2}
\end{figure}

\begin{figure}[!btp]
\centering
\includegraphics[trim = 0cm 0cm 0cm -0.5cm,width = 0.45\textwidth]{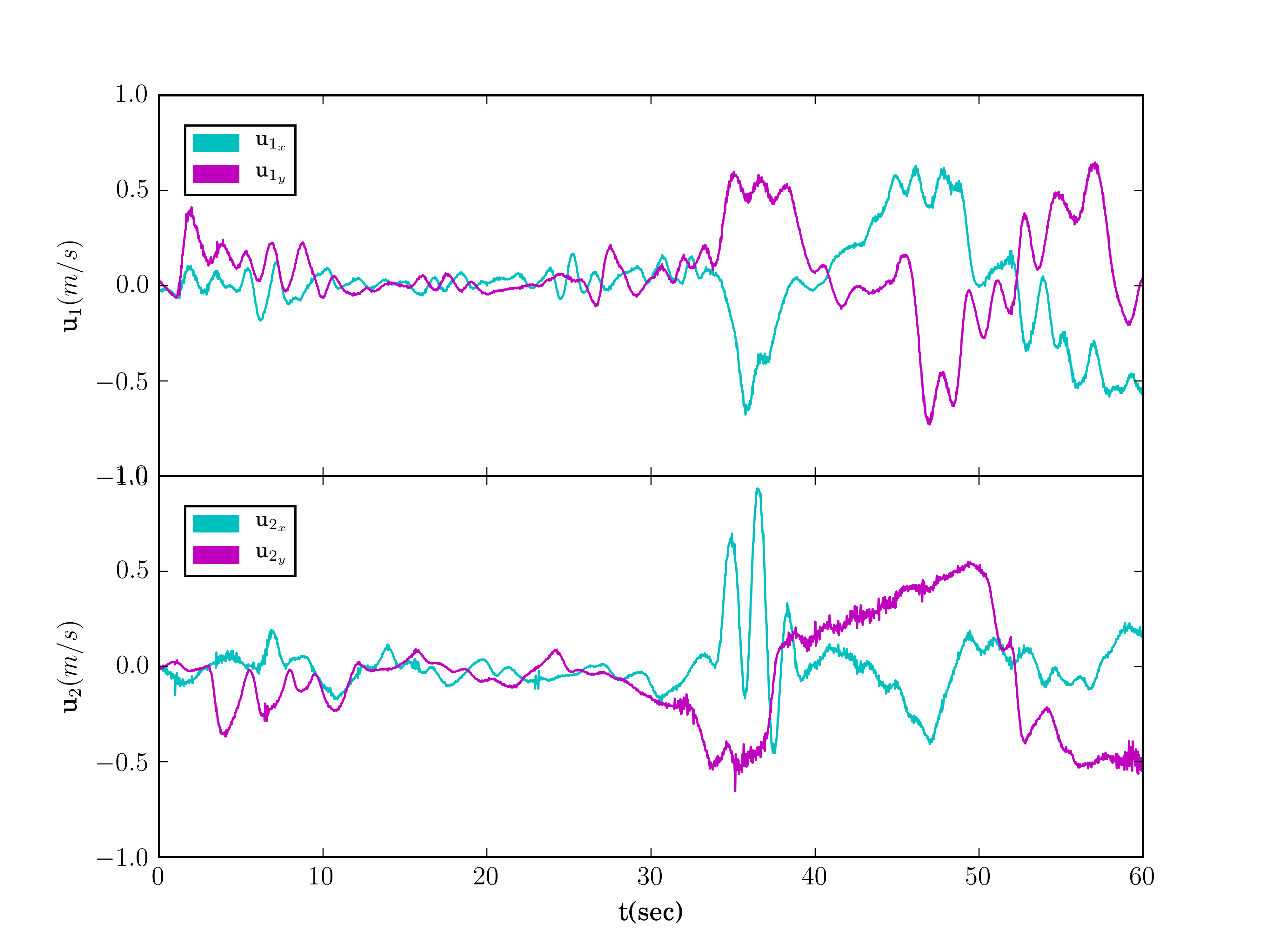}

\caption{The resulting $2$-dimensional control signals of the $2$ agents for the second experimental scenario. Top: agent 1, bottom: agent 2. \label{fig:exp_2_vel}}
\end{figure}

\begin{figure*}[!tbp]
  \centering
  \subfloat[]{\includegraphics[trim = 0cm 0cm 0cm -0.75cm,width=0.25\textwidth,height=0.19\textwidth]{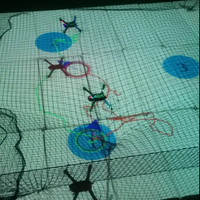}\label{fig:exp_2_path:1}}
  \quad
  \subfloat[]{\includegraphics[trim = 0cm 0cm 0cm -0.75cm,width=0.25\textwidth,height=0.19\textwidth]{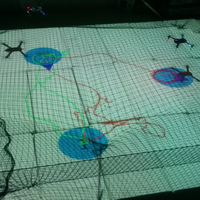}\label{fig:exp_2_path:2}}
  \quad
  \subfloat[]{\includegraphics[trim = 0cm 0cm 0cm -0.75cm,width=0.25\textwidth,height=0.19\textwidth]{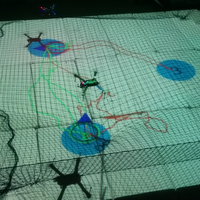}\label{fig:exp_2_path:3}}
  \hfill
  \subfloat[]{\includegraphics[trim = 1cm 0cm 0cm 0cm,width=0.25\textwidth, height = 0.2\textwidth]{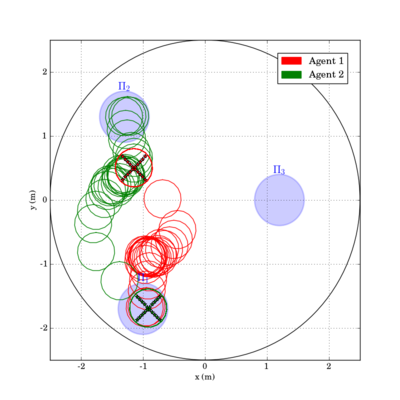}\label{fig:exp_2_path:1_sim}}
  \quad	
  \subfloat[]{\includegraphics[trim = 1cm 0cm 0cm 0cm,width=0.25\textwidth, height = 0.2\textwidth]{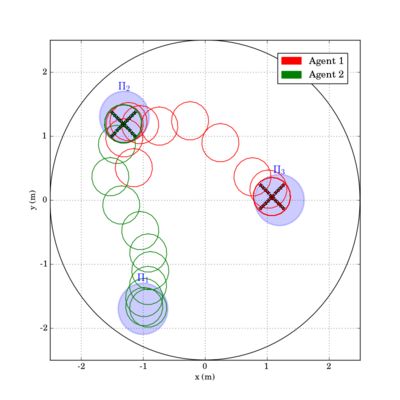}\label{fig:exp_2_path:2_sim}}
  \quad
  \subfloat[]{\includegraphics[trim = 1cm 0cm 0cm 0cm,width=0.25\textwidth, height = 0.2\textwidth]{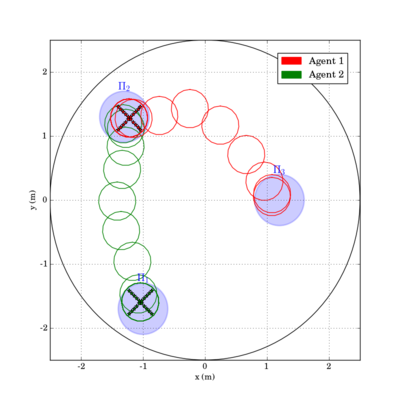}\label{fig:exp_2_path:3_sim}}
  \caption{Execution of the paths $(\pi_1\pi_2\pi_3\pi_2)^1$ and $(\pi_2\pi_1)^2$ by agents $1$ and $2$, respectively for the second experimental scenario. (a), (d): $\pi_1\rightarrow_1\pi_2, \pi_2\rightarrow_2\pi_1$, (b), (e): $\pi_2\rightarrow_1\pi_3, \pi_1\rightarrow_2\pi_2$, (c), (f): $\pi_3\rightarrow_1\pi_2, \pi_2\rightarrow_2\pi_1$.}\label{fig:exp_2_path}
\end{figure*}

The second experimental scenario included $3$ regions of interest $\Pi=\{\pi_1,\dots,\pi_3\}$ in $\mathcal{W}$, with $r_{\pi_k} = 0.4,\forall k\in\{1,\dots,3\}$ and $\boldsymbol{p_{\pi_1}} = [-1,-1.7]^T$m, $\boldsymbol{p_{\pi_2}} = [-1.3,1.3]^T$m and $\boldsymbol{p_{\pi_3}} = [1.2,0]^T$m. The initial positions of the agents were taken such that $\mathcal{A}_1(\boldsymbol{p_1}(0)) \in\pi_1$ and $\mathcal{A}_i(\boldsymbol{p_2}(0)) \in\pi_2$ (see Fig. \ref{fig:Initial_workspace_exp_2}). We also defined the atomic propositions $\Psi_1 = \Psi_2 = \{``\text{res}_\text{a}",``\text{res}_\text{b}",``\text{base}"\}$, corresponding to a base and several resources in the workspace, with $L_1(\pi_1) = L_2(\pi_1)=\{``\text{res}_\text{a}"\}, L_1(\pi_2) = L_2(\pi_2)= \{``\text{base}"\}, L_1(\pi_3) = L_2(\pi_3)=\{``\text{res}_\text{b}"\}$. We considered that the agents had to transfer the resources to the "base" in $\pi_2$; both agents were responsible for $``\text{res}_\text{a}"$ but only agent $1$ should access $``\text{res}_\text{b}"$. The specifications were translated to the formulas $\phi_1 = \square(\lozenge(``\text{res}_\text{a}"\bigcirc``\text{base}")\land\lozenge(``\text{res}_\text{b}"\bigcirc``\text{base}")), \phi_2 = \square\neg``\text{res}_\text{b}"\land\square\lozenge(``\text{res}_\text{a}"\bigcirc``\text{base}")$ and the derived paths were $p_1 = (\pi_1\pi_2\pi_3\pi_2)^\omega$ and $p_2 = (\pi_1\pi_2)^\omega$. The execution of the paths $(\pi_1\pi_2\pi_3\pi_2)^1$ and $(\pi_2\pi_1)^2$ by agents 1 and 2, respectively, are depicted in Fig. \ref{fig:exp_2_path}, and the corresponding control inputs are shown in Fig. \ref{fig:exp_2_vel}. The figures demonstrate the successful execution and satisfaction of the paths and formulas, respectively, and the compliance with the control input bounds. 

Regarding the continuous control protocol in the aforementioned experiments, we chose $k_{g_i} = 3, \lambda_i = 2$ in (\ref{eq:feedback_contr}), (\ref{eq:feedback_contr_2}) and the switching duration in (\ref{eq:switch_controller}) as $\nu_i = 0.1t'_{i,k}, \forall i\in\{1,2\}$.


\begin{rem}
Note that, although the limited available workspace in the experiments did not satisfy all the conditions regarding the distance between regions and the workspace boundary, as introduced in Section \ref{sec:System and PF}, the two experimental scenarios were successfully conducted.
\end{rem}

The simulations and experiments were conducted in Python environment using an Intel Core i7 2.4 GHz personal computer with 4 GB of RAM, and are clearly demonstrated in the video  found in https://youtu.be/dO77ZYEFHlE, a compressed version of which has been submitted with this paper.

\section{Conclusion and Future Work} \label{sec:conclusion}
In this work, we proposed a control strategy for the motion planning of a team of aerial vehicles under LTL specifications. By using decentralized navigation functions that guarantee inter-agent collision avoidance, we abstracted each agent's motion as a finite transition system between regions of interest. Each agent then derived the plan that satisfies its given LTL formula through formal-verification techniques. Simulation studies and experimental results verified the efficiency of the proposed algorithm. Future efforts will be devoted towards considering more complex, second order dynamics, partially known environments and experiments with more agents. 

\bibliographystyle{IEEEtran}
\bibliography{bibliography}

\end{document}